\documentclass{bmvc2k}

\def\paperTitle{FADE: Few-shot/zero-shot Anomaly Detection Engine using Large Vision-Language Model}
\def\paperTitleShort{Few-shot/zero-shot Anomaly Detection Engine}

\def\authorBlock{
    \addauthor{Yuanwei Li\footnotemark[1]\footnotemark[2]}{yuanwei_li@hotmail.com}{1}
    \addauthor{Elizaveta Ivanova\footnotemark[1]}{lisa.ivanova@entrust.com}{1}
    \addauthor{Martins Bruveris}{martins.bruveris@entrust.com}{1}

    \addinstitution{
     Onfido\\
     London, UK
    }
}

\def\runningHEAD{
Li \etal
}

\def\etal{\emph{et al}\bmvaOneDot}
\PassOptionsToPackage{dvipsnames}{xcolor}
\usepackage{booktabs}
\usepackage{pifont}
\usepackage{caption}
\usepackage{subcaption}
\usepackage{amssymb}
\usepackage{makecell}
\usepackage{graphicx} 
\usepackage{enumitem}
\usepackage[export]{adjustbox}
\usepackage{wrapfig}
\usepackage{xcolor}
\newcommand{\mysubsubsection}[1]{\vspace{0.07cm} \noindent {\bf #1}:}

\title{\paperTitle}

\authorBlock

\runninghead{\runningHEAD}{\paperTitleShort}

\begin{document}

\footnotetext[1]{These authors contributed equally.}
\footnotetext[2]{Corresponding author.}
\maketitle

\begin{abstract}
Automatic image anomaly detection is important for quality inspection in the manufacturing industry. The usual unsupervised anomaly detection approach is to train a model for each object class using a dataset of normal samples. However, a more realistic problem is zero-/few-shot anomaly detection where zero or only a few normal samples are available. This makes the training of object-specific models challenging. Recently, large foundation vision-language models have shown strong zero-shot performance in various downstream tasks. While these models have learned complex relationships between vision and language, they are not specifically designed for the tasks of anomaly detection. In this paper, we propose the Few-shot/zero-shot Anomaly Detection Engine (FADE) which leverages the vision-language CLIP model and adjusts it for the purpose of industrial anomaly detection. Specifically,  we improve language-guided anomaly segmentation 1) by adapting CLIP to extract multi-scale image patch embeddings that are better aligned with language and 2) by automatically generating an ensemble of text prompts related to industrial anomaly detection. 3) We use additional vision-based guidance from the query and reference images to further improve both zero-shot and few-shot anomaly detection. On the MVTec-AD (and VisA) dataset, FADE outperforms other state-of-the-art methods in anomaly segmentation with pixel-AUROC of 89.6\% (91.5\%) in zero-shot and 95.4\% (97.5\%) in 1-normal-shot. Code is available at \href{https://github.com/BMVC-FADE/BMVC-FADE}{https://github.com/BMVC-FADE/BMVC-FADE}.
\end{abstract}
\section{Introduction}
\label{sec:intro}
\begin{figure}[h!]
\centering
\begin{minipage}[t]{0.52465\linewidth}
\centering
\includegraphics[valign=c,width=\textwidth]{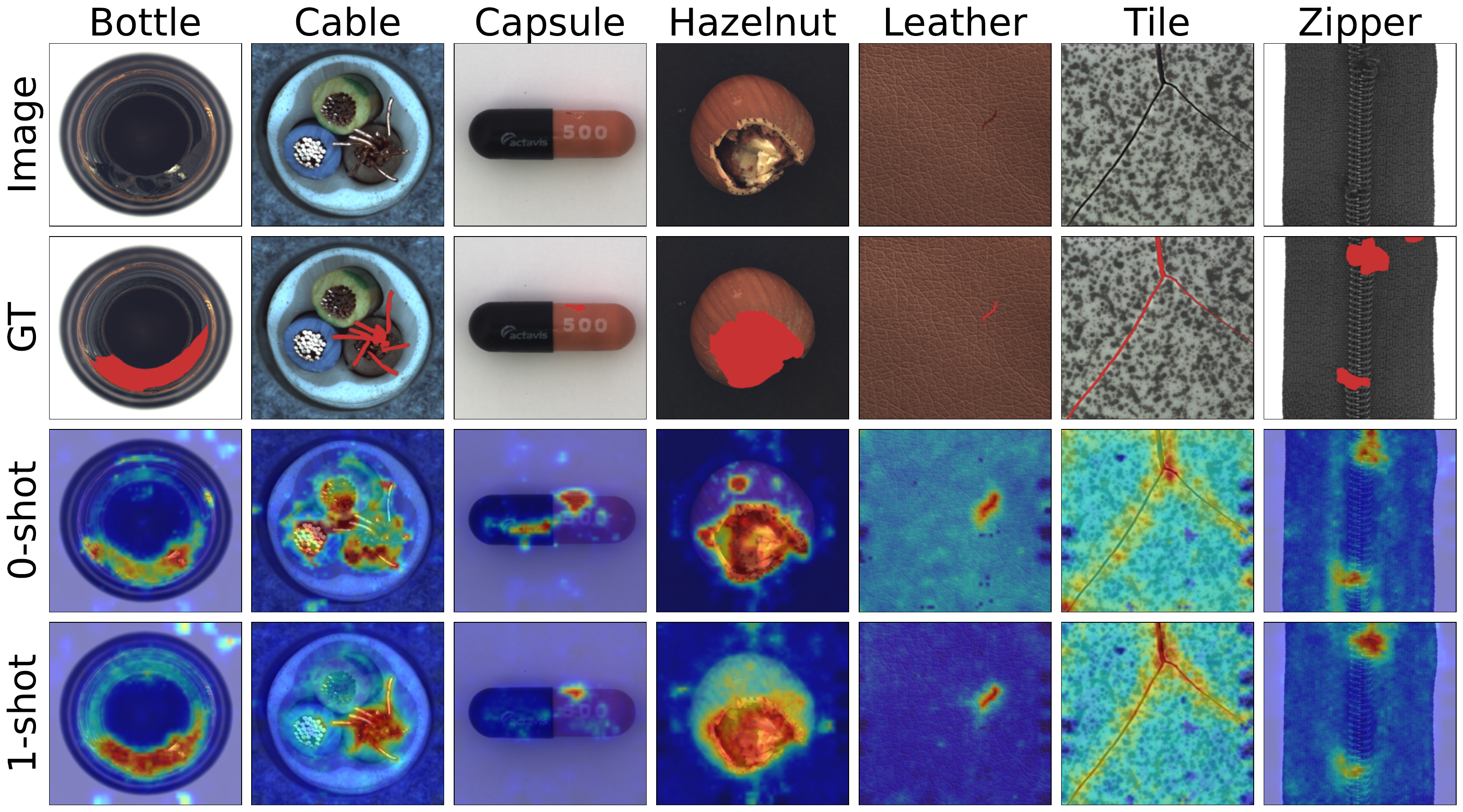}
\captionsetup{labelformat=empty,font=small,skip=1pt}
\caption*{(a) MVTec-AD}\label{fig:qual_mvtec}
\end{minipage}
\hspace{0.000001\linewidth}
\begin{minipage}[t]{0.41475\linewidth}
\centering
\includegraphics[valign=c,width=\textwidth]{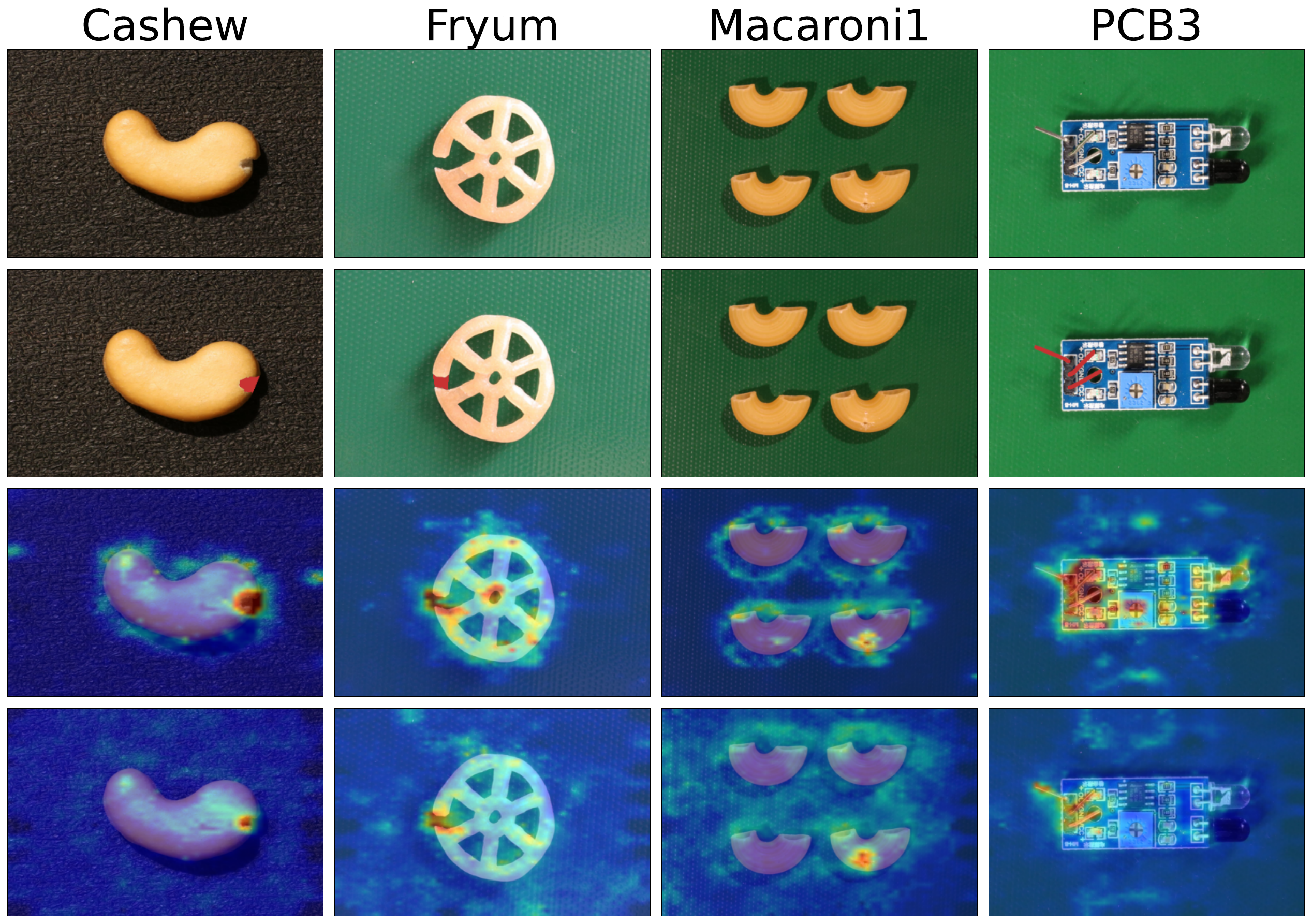}
\captionsetup{labelformat=empty,font=small,skip=1pt}
\caption*{(b) VisA}\label{fig:qual_visa}
\end{minipage}%
\begin{minipage}[t]{0.048\linewidth}
\centering
\includegraphics[valign=c,width=\textwidth]{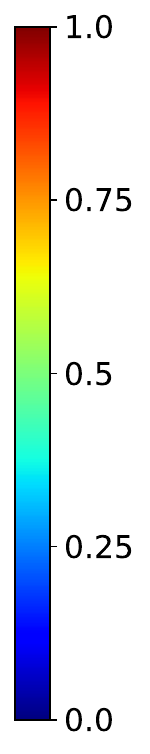}
\captionsetup{labelformat=empty,type=figure}
\caption*{}
\end{minipage}

\caption{Zero-/one-shot anomaly segmentation results of FADE.}
\label{fig:qualitative_main}
\end{figure}

Industrial image anomaly detection involves anomaly classification (AC) and anomaly segmentation (AS) which aim to identify and localise anomalies occurring on objects found in manufacturing industries. Most of the current research focuses on unsupervised anomaly detection where only normal samples are used during training while both normal and anomalous samples are evaluated during inference~\cite{defard2021padim, roth2022patchcore, rudolph2021differnet, gudovskiy2022cflow, bergmann2020studentteacher, li2021cutpaste}. Some of these approaches are able to achieve high performance when a large number of normal training samples is available. But this is not a realistic setting for industrial anomaly detection for two reasons. First, the number of normal training samples can be scarce or unavailable. Second, one specific model needs to be trained for each object class and this quickly becomes unscalable when the number of object classes increases.

Zero-shot/few-normal-shot anomaly detection is a more realistic setting with either zero or only a few normal reference images available. However, many previous methods designed for unsupervised anomaly detection perform poorly under this low-data regime. There are increasingly more research works focusing on this area recently~\cite{wu2021metaformer, huang2022regad, jeong2023winclip, cao2023saa, zhou2024anomalyclip} but there is still much room for improvement.

Vision foundation models trained on large image datasets have recently shown superb zero-shot capability on various computer vision tasks such as image classification, object detection and segmentation~\cite{radford2021clip, liu2023groundingdino, kirillov2023sam}. There is great potential to leverage and transfer the knowledge and representations learned in these foundation models for the zero-/few-shot anomaly detection task. But these foundation models are not specifically designed and trained for anomaly detection. As such, we propose the Few-shot/zero-shot Anomaly Detection Engine (FADE) that utilises one of these foundation models, the CLIP model~\cite{radford2021clip}, and adapt it for the purpose of industrial AC and AS.

CLIP is a visual-language model pretrained on large datasets of image-text pairs using a contrastive loss. CLIP learns concepts and representations that capture the relationship between language and image. Without any finetuning, the model has demonstrated zero-shot capability in various downstream tasks such as image classification. This is done by constructing free-form text prompts as classification labels which enable the use of natural language to extract relevant information already learned by the CLIP model. This allows CLIP to be a suitable candidate for handling the anomaly detection task. The concept of anomaly and normality is broad and abstract. But language allows us to describe a specific anomaly precisely (E.g. a cracked object, a scratched surface, a missing component). CLIP can leverage the power of language to better capture the concept of anomaly and hence improve upon zero-shot anomaly detection performance. However, this requires careful prompt engineering and some methods~\cite{jeong2023winclip} manually design an ensemble of text prompts with key words such as ``damage'' and ``defect'' which can be a time-consuming process. As such, we propose to utilise existing Large Language Model (LLM) to automatically generate text prompts related to the concept of normality and anomaly.

There is another challenge associated with language-guide anomaly detection. ViT-based CLIP model is trained by aligning the image-level CLS token embeddings with the text embeddings. This works for image-level anomaly classification but not pixel-level anomaly segmentation. This is because language-guided segmentation requires the comparison between the image patch embeddings and the text embeddings. However, the patch embeddings are not aligned to natural language during training. This results in sub-optimal segmentation performance~\cite{li2023clipsurgery, bousselham2023gem, jeong2023winclip}. To address this issue, we propose to apply the Grounding Everything Module (GEM)~\cite{bousselham2023gem} to extract the image patch embeddings which are shown to have better alignment with language and perform better in zero-shot segmentation.

While CLIP can be used for language-driven anomaly detection, we can also only use its image encoder for vision-based anomaly detection. In this case, visual representations of image patches extracted from both query and reference images are compared to one another to identify inconsistencies and anomalies. This enhances zero-shot performance and extends the approach to the few-shot setting when normal reference images are available.

In summary, we propose FADE which uses pretrained CLIP model for zero-/few-shot anomaly detection without any further training or fine-tuning. Our main contributions are:
\begin{itemize}[itemsep=2pt]
    \item We utilise GEM patch embeddings which are better aligned with language to improve zero-shot language-guided AS. In addition, we adopt a multi-scale approach to make the method robust at detecting anomalies of different sizes.
    \item We further improve language-guided anomaly detection by using an LLM to automatically generate a prompt ensemble that captures the concept of normality and anomaly using a diverse set of text prompts which are related to industrial anomaly inspection.
    \item We enhance anomaly detection performance by employing a vision-guided approach that can be applied to both zero- and few-shot settings.
    \item FADE shows competitive results on the MVTec-AD and VisA benchmarks for AC and AS under both the zero-shot and few-normal-shot regimes.
\end{itemize}
\section{Related Work}
\label{sec:related}

\mysubsubsection{Anomaly detection}
Most research has focused on unsupervised anomaly detection where a model is trained on many normal training samples~\cite{defard2021padim, roth2022patchcore, rudolph2021differnet, gudovskiy2022cflow, bergmann2020studentteacher, li2021cutpaste}. Among them, PatchCore~\cite{roth2022patchcore}, a memory bank based method, has achieved state-of-the-art performance. There has been a growing interest in zero-/few-shot anomaly detection since the performance of the above methods decreases under such regimes. Earlier work such as Metaformer~\cite{wu2021metaformer} addresses the problem in a meta-learning framework. RegAD~\cite{huang2022regad} detects anomalies by comparing registered features of a test image with the normal reference images. However, both methods require model training using some datasets. A recent work, Segment Any Anomaly (SAA)~\cite{cao2023saa}, leverages a foundation segmentation model SAM~\cite{kirillov2023sam} for zero-shot AS.

\mysubsubsection{Anomaly detection with CLIP}
Recently, CLIP has shown impressive language-driven zero-shot capability in various computer vision tasks~\cite{radford2021clip}. WinCLIP~\cite{jeong2023winclip} is one of the first to use CLIP for language-guided zero-/few-shot anomaly detection. The paper adapts CLIP for pixel-wise AS by extracting dense CLIP features based on overlapping windows. Subsequent papers APRIL-GAN~\cite{chen2023aprilgan} and AnomalyCLIP~\cite{zhou2024anomalyclip} build upon the general framework of WinCLIP and introduce additional learnable layers and learnable prompts respectively to fine-tune the model for AS. ClipSAM~\cite{li2024clipsam} leverages CLIP to obtain a rough anomaly segmentation mask from which points and bounding boxes are then extracted and used by SAM as input prompts to further refine the segmentation. However, all the above methods use auxiliary datasets with ground truth anomaly segmentation masks for training. Other methods explore prompting in detail. \cite{li2024promptad} constructs manual and learnable normal and anomaly prompts and proposes a one-class prompt-learning method for few-shot detection. \cite{tamura2023randomword} augments normal and anomalous text prompts by inserting random words and uses their CLIP text embeddings to train a feed-forward network for zero-shot AC. Our method, unlike most of the above works, does not require further training or fine-tuning.

\mysubsubsection{Zero-shot segmentation with CLIP}
While CLIP shows strong zero-shot image classification capability, there are challenges to adapt it for dense pixel-wise segmentation. The main problem is that CLIP training does not directly optimise for alignment between language and local image patch embeddings~\cite{jeong2023winclip}. In addition, empirical segmentation results show that CLIP tends to generate opposite visualisation between foreground/background and also produce noisy activations~\cite{li2023clipsurgery}. Several works have attempted to tackle these issues by adapting the CLIP model architecture for zero-shot segmentation~\cite{zhou2022maskclip, li2023clipsurgery, bousselham2023gem}. CLIP Surgery~\cite{li2023clipsurgery} replaces the query-key attention in the original transformer block with a value-value attention and also removes the feed-forward network. \cite{bousselham2023gem} proposes GEM blocks which further generalises the idea to self-self attention. In our work, we apply the GEM blocks to the task of anomaly segmentation.
\section{Method}
\label{sec:method}

FADE conducts both anomaly classification (AC) and anomaly segmentation (AS) under the zero-/few-shot settings. It consists of four different anomaly detection pipelines that are used under different cases (Fig.~\ref{fig:method}a-d). The detection pipeline can be either language- or vision-guided. Language-guided detection utilises the language capability of the CLIP model to perform zero-shot AC (Fig.~\ref{fig:method}a) and AS (Fig.~\ref{fig:method}b). On the other hand, vision-guided detection employs visual cues from images to further improve zero-shot AS (Fig.~\ref{fig:method}c) and extends the approach to few-shot detection (Fig.~\ref{fig:method}d).

\subsection{Language-Guided Anomaly Classification}
\label{subsec:lang_ac}
\begin{figure*}[tp]
    \centering
    \includegraphics[width=\linewidth]{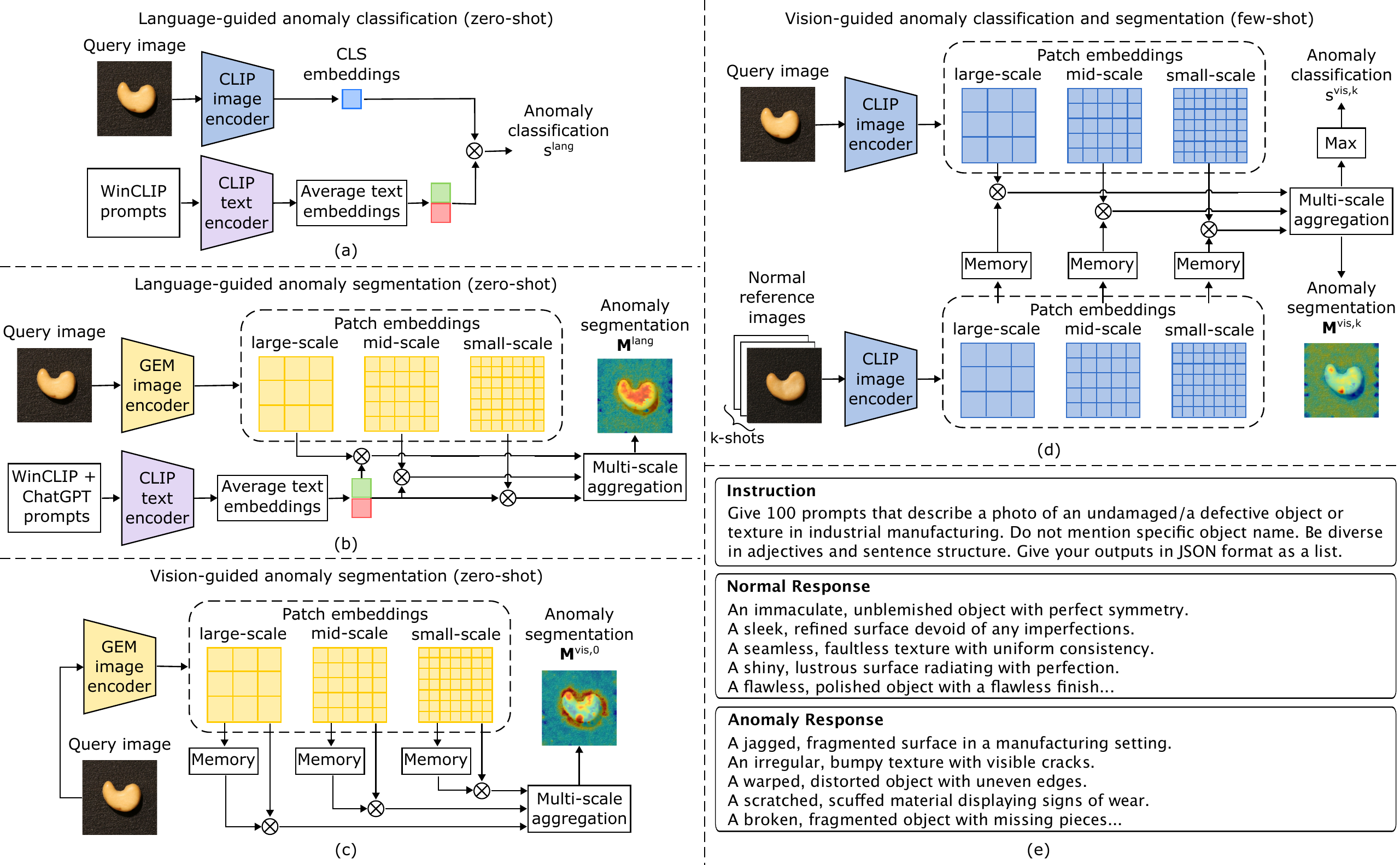}
    \caption{Different components of FADE. (a) Zero-shot language-guided AC; (b) Zero-shot language-guided AS; (c) Zero-shot vision-guided AS; (d) Few-shot vision-guided AC and AS; (e) ChatGPT prompts generation: An instruction given to ChatGPT and some of its responses.}
    \label{fig:method}
\end{figure*}

Zero-shot language-guided AC (Fig.~\ref{fig:method}a) is carried out as described in WinCLIP~\cite{jeong2023winclip}. A prompt ensemble $\boldsymbol{t}=\{ \boldsymbol{t_{+}}, \boldsymbol{t_{-}} \}$ is manually crafted which captures the notion of object anomaly or normality (Appendix A of \cite{jeong2023winclip}) where $\boldsymbol{t_{+}}$ and $\boldsymbol{t_{-}}$ are the sets of 88 anomalous and 154 normal text prompts respectively. An example of a normal prompt is ``\texttt{a cropped photo of the flawless [o]}'' while an anomalous prompt can be ``\texttt{a cropped photo of the damaged [o]}'' where \texttt{[o]} is the object label that comes with the dataset (E.g. bottle). The CLIP text encoder $g$ extracts the text embeddings of all the prompts. The average text embeddings $\boldsymbol{h}$ for the anomalous and normal text prompts are computed as $\boldsymbol{h_{+}} = \frac{1}{N_{\boldsymbol{t_{+}}}} {}\sum_{t \in \boldsymbol{t_{+}}}^{} g(t)$ and $\boldsymbol{h_{-}} = \frac{1}{N_{\boldsymbol{t_{-}}}} {}\sum_{t \in \boldsymbol{t_{-}}}^{} g(t)$ where $N_{\boldsymbol{t_{+}}}$ and $N_{\boldsymbol{t_{-}}}$ are the number of anomalous and normal prompts respectively. Given an image $\boldsymbol{x}$ and the CLIP image encoder $f^{clip}$, we compute its image embeddings $f_{cls}^{clip}(\boldsymbol{x})$ where $cls$ indicates the CLS token of the ViT-based CLIP model. The cosine similarity $\langle f_{cls}^{clip}(\boldsymbol{x}), \boldsymbol{h} \rangle$ then gives a measure of how close the image is to the concept of normality or anomaly as encapsulated by the text prompts. Binary zero-shot AC is performed by calculating a language-guided anomaly score:
\begin{equation}  \label{eq:zeroshot_ac}
s^{lang}=\frac{\exp\big(\langle f_{cls}^{clip}(\boldsymbol{x}), \boldsymbol{h_{+}} \rangle /\tau \big)}{\exp\big(\langle f_{cls}^{clip}(\boldsymbol{x}), \boldsymbol{h_{+}} \rangle /\tau \big) + \exp\big(\langle f_{cls}^{clip}(\boldsymbol{x}), \boldsymbol{h_{-}} \rangle /\tau \big)}
\end{equation}
where $\tau$ is a temperature parameter fixed at 0.01.

\subsection{Language-Guided Anomaly Segmentation}
\label{subsec:lang_as}

\mysubsubsection{GEM embeddings}
Given an image of size $h\times w$, AS generates a dense pixel-level prediction $\boldsymbol{M} \in [0, 1]^{h\times w}$ which localises the anomalous regions. The above framework for AC can be extended to zero-shot language-guided AS (Fig.~\ref{fig:method}b). Specifically, the patch embeddings $\{f_{p}^{clip}(\boldsymbol{x})\}_{p \in Patches}$ from the last transformer block of the CLIP image encoder are extracted to replace the CLS embeddings $f_{cls}^{clip}(\boldsymbol{x})$ in Eq.~\ref{eq:zeroshot_ac}, where $Patches$ is a set of all patches extracted from an image $x$. This enables anomaly prediction at each location of a patch $p$ that can be spatially combined to form a dense segmentation map. However, this approach yields poor results~\cite{jeong2023winclip,bousselham2023gem} since CLIP is trained using an image-level contrastive loss that aligns the image-level CLS embeddings with the text embeddings. Hence, CLIP has poor alignment between its patch embeddings and the text embeddings, resulting in its failure to generalise its zero-shot capability to the dense segmentation task. WinCLIP~\cite{jeong2023winclip} solves this issue by introducing a grid of overlapping windows where each window masks out the image content outside of the window and computes the CLS embeddings of the content inside the window. This produces a dense segmentation while maintaining the visual-language alignment. 

\begin{sloppypar}
We take a different approach based on the Grounding Everything Module (GEM)~\cite{bousselham2023gem} that has demonstrated success in zero-shot open-vocabulary object localisation. GEM uses the idea of self-self attention which computes query-query, key-key and value-value attention weights instead of the query-key attention weights in the conventional self-attention mechanism. Mathematically, self-self attention weights are given by: 
\begin{math}
{Attn_{self\text{-} self} = \operatorname{softmax}\big(\boldsymbol{h}_{Patches}^{clip}W\cdot (\boldsymbol{h}_{Patches}^{clip}W)^T\big)}
\end{math} 
where $\boldsymbol{h}_{Patches}^{clip}\in \mathbb{R}^{N\times d}$ are the patch embeddings of dimension $d$ from the previous transformer block of the original CLIP model for all $N$ patches in an image. $W \in{\{ W_{v}, W_{q}, W_{k} \}}$ are the projection matrices of the current transformer block. The outputs from q-q, k-k and v-v attention are ensembled together to form the output of a GEM block. The GEM block also removes the feed-forward network from the conventional transformer block and is constructed as a parallel pathway to the original transformer blocks pathway. The outputs of all the GEM blocks at different layers are summed together to obtain the final GEM patch embeddings $\{f_{p}^{gem}(\boldsymbol{x})\}_{p \in Patches}$. The zero-shot language-guided AS mask $\boldsymbol{M}^{lang}=\big\{M_p^{lang}\big\}_{p \in Patches}$ is then obtained where each patch anomaly score $M_p^{lang}$ is computed as:
\begin{equation} \label{eq:lang_gem_def}
M_p^{lang} = \frac{\exp\big(\langle f_{p}^{gem}(\boldsymbol{x}), \boldsymbol{h_{+}} \rangle /\tau \big)}{\exp\big(\langle f_{p}^{gem}(\boldsymbol{x}), \boldsymbol{h_{+}} \rangle /\tau \big) + \exp\big(\langle f_{p}^{gem}(\boldsymbol{x}), \boldsymbol{h_{-}} \rangle /\tau \big)}
\end{equation}.

GEM embeddings have shown better visual-language alignment without further fine-tuning the CLIP model. It also has a connection to clustering properties that group similar pixels together, resulting in better zero-shot segmentation~\cite{bousselham2023gem}. GEM patch embeddings are computed based on the global context of the entire image which is important for anomaly segmentation (E.g. anomalous regions are better identified through comparison with normal regions). In contrast, WinCLIP embeddings only have a limited context restricted to a local window. In addition, GEM can be computationally more efficient with a single forward pass of the whole image.
\end{sloppypar}

\mysubsubsection{Prompt engineering}
CLIP is pretrained on large datasets of visual-language pairs and learns powerful representations and alignment between images and texts that are useful for anomaly detection. Prompt engineering that captures the specific description and concept of anomaly is essential for maximising the zero-shot capability of the CLIP model. While the WinCLIP prompt ensemble has been designed manually to optimise for anomaly detection, we notice that AS performance can be further improved with additional prompt engineering.

The concept of anomaly and normality is broad and goes beyond the limited descriptions provided by the WinCLIP prompts that use mostly general words like ``damage'', ``flaw''. In contrast, the MVTec and VisA datasets contain different types of specific industrial anomalies such as ``a scratch'', ``a crack'', ``a missing part''. While these specific descriptions are useful for AS, manually crafting them is time-consuming and unscalable. Instead, we use a large language model, ChatGPT 3.5\footnote{https://chat.openai.com}, to automatically generate a large number of text prompts that contain a diverse set of specific descriptions related to industrial anomaly detection. Fig.~\ref{fig:method}e shows an instruction prompt that is passed to ChatGPT 3.5 and some of its responses. The instruction ensures that specific object names are not mentioned in the response so that the generated text prompts are object-agnostic. This is because the notion of the specific object becomes less relevant for AS when the patch embeddings focus on local anomalous regions that are only part of the object. We also attempt to add information regarding the anomaly size and location in the text prompts but it did not improve the results.

We use 5 instruction prompts with different wordings to obtain responses that are more diverse and varied. This generates a total of 486 anomaly and 423 normal prompts that form the ChatGPT prompt ensemble. The instructions and their responses are listed in supplementary. The ChatGPT and WinCLIP prompt ensembles are combined and used for AS.

\mysubsubsection{Multi-scale aggregation}
We extract multi-scale GEM embeddings in order to detect anomalies with different sizes. Although the CLIP model is trained at a fixed image size, it can take inputs of different image sizes during inference by interpolating the positional embeddings. We resize the original image to 3 input sizes: $240\times 240$, $448\times 448$, $896\times 896$. Given that the patch size of the model is 16, this corresponds to output segmentations with size $15\times 15$, $28\times 28$ and $56\times 56$. The segmentations are then upscaled to a fixed size and averaged to give the final segmentation map $\boldsymbol{M}^{lang}$.

\subsection{Vision-Guided Anomaly Classification and Segmentation}
\label{subsec:vision_ac_as}

Vision-guided anomaly detection extracts and compares visual cues from images in order to identify anomalies. This is complementary to the language-guided approach and further extends it to the few-shot setting.

\mysubsubsection{Few-shot setting}
The performance of AC and AS improves when a few reference normal images are available. Specifically, we follow the approach similar to WinCLIP~\cite{jeong2023winclip} where we build a memory bank $\boldsymbol{R}$ of CLIP patch embeddings extracted from the $k$ reference images (Fig.~\ref{fig:method}d). We have also tried to use GEM embeddings but the results are inferior. Given a query image $\boldsymbol{x}$ and its CLIP embeddings $f_{p}^{clip}(\boldsymbol{x})$ for a patch $p$, the vision-guided anomaly score for this patch under the $k$-shot setting is computed as the cosine distance to its nearest neighbour patch in the memory bank:
\begin{math}
M_{p}^{vis,k}=\min_{r\in \boldsymbol{R}} \frac{1}{2}\Big(1-\big\langle f_{p}^{clip}(\boldsymbol{x}), r \big\rangle\Big)
\end{math}.
Spatially combining the anomaly scores of the patches at all locations in an image gives the vision-guided segmentation map. We can extend this to a multi-scale approach by building a separate memory bank for the patch embeddings extracted from the reference images at different scales (E.g.~different image sizes). The multi-scale segmentation maps are averaged to give the final vision-guided segmentation $\boldsymbol{M}^{vis,k}$ under the k-shot setting. In addition, we use the maximum value of $\boldsymbol{M}^{vis,k}$ as an anomaly score $s^{vis,k}$ for few-shot vision-guided AC.

\mysubsubsection{Zero-shot setting}
Vision-guided AS can also be applied in the zero-shot setting (Fig.~\ref{fig:method}c) with 2 differences compared to the few-shot setting: 1) The memory bank is built using patches from the query image itself since reference images are unavailable. In this case, the resulting anomaly score is computed as the distance to the second nearest neighbour patch since the nearest neighbour is the patch itself. 2) GEM patch embeddings are used since they perform better than CLIP patch embeddings. Multi-scale aggregation is used again to give the final segmentation $\boldsymbol{M}^{vis,0}$ under the zero-shot setting.

\mysubsubsection{Combining language and vision guidance}
Language and vision guidance can be combined to improve overall AC and AS performance. Tab.~\ref{tab:lang_vis_aggregation} summarises how the combination is done under the different settings. Refer to supplementary for details.
\section{Experiments}
\label{sec:experiments}
\begin{table}[t]
    \centering
    \begin{minipage}[c]{0.695\textwidth}
        \centering
        \resizebox{\textwidth}{!}{%
            \begin{tabular}{clcccccc}
                \toprule
                \multicolumn{2}{l}{Anomaly Classification} & \multicolumn{3}{c}{MVTec-AD} & \multicolumn{3}{c}{VisA} \\
                \cmidrule(lr){3-5} \cmidrule(lr){6-8}
                Setup & Method & AUROC & AUPR & $F1$-max & AUROC & AUPR & $F1$-max\\
                \midrule
                0-shot & WinCLIP & \textbf{91.8$\pm$\small{0.0}} & \textbf{96.5$\pm$\small{0.0}} & \textbf{92.9$\pm$\small{0.0}} & \textbf{78.1$\pm$\small{0.0}} & \textbf{81.2$\pm$\small{0.0}} & \textbf{79.0$\pm$\small{0.0}} \\
                \cmidrule(lr){2-8}
                & FADE (ours) & 90.0$\pm$\small{0.0} & 95.6$\pm$\small{0.0} & 92.4$\pm$\small{0.0} & 75.6$\pm$\small{0.0} & 78.5$\pm$\small{0.0} & 78.6$\pm$\small{0.0} \\
                \midrule
                1-shot & PatchCore & 83.4$\pm$\small{3.0} & 92.2$\pm$\small{1.5} & 90.5$\pm$\small{1.5} & 79.9$\pm$\small{2.9} & 82.8$\pm$\small{2.3} & 81.7$\pm$\small{1.6} \\
                & WinCLIP+ & 93.1$\pm$\small{2.0} & 96.5$\pm$\small{0.9} & 93.7$\pm$\small{1.1} & 83.8$\pm$\small{4.0} & 85.1$\pm$\small{4.0} & 83.1$\pm$\small{1.7} \\
                \cmidrule(lr){2-8}
                & FADE (ours) & \textbf{93.9$\pm$\small{0.7}} & \textbf{96.8$\pm$\small{0.3}} & \textbf{94.8$\pm$\small{0.2}} & \textbf{86.7$\pm$\small{2.0}} & \textbf{87.9$\pm$\small{1.5}} & \textbf{84.7$\pm$\small{0.8}} \\
                \midrule
                2-shot & PatchCore & 86.3$\pm$\small{3.3} & 93.8$\pm$\small{1.7} & 92.0$\pm$\small{1.5} & 81.6$\pm$\small{4.0} & 84.8$\pm$\small{3.2} & 82.5$\pm$\small{1.8} \\
                & WinCLIP+ & 94.4$\pm$\small{1.3} & 97.0$\pm$\small{0.7} & 94.4$\pm$\small{0.8} & 84.6$\pm$\small{2.4} & 85.8$\pm$\small{2.7} & 83.0$\pm$\small{1.4} \\
                \cmidrule(lr){2-8}
                & FADE (ours) & \textbf{95.2$\pm$\small{1.0}} & \textbf{97.6$\pm$\small{0.5}} & \textbf{95.0$\pm$\small{0.4}} & \textbf{89.2$\pm$\small{0.4}} & \textbf{90.2$\pm$\small{0.2}} & \textbf{85.9$\pm$\small{0.6}} \\
                \midrule
                4-shot & PatchCore & 88.8$\pm$\small{2.6} & 94.5$\pm$\small{1.5} & 92.6$\pm$\small{1.6} & 85.3$\pm$\small{2.1} & 87.5$\pm$\small{2.1} & 84.3$\pm$\small{1.3} \\
                & WinCLIP+ & 95.2$\pm$\small{1.3} & 97.3$\pm$\small{0.6} & 94.7$\pm$\small{0.8} & 87.3$\pm$\small{1.8} & 88.8$\pm$\small{1.8} & 84.2$\pm$1.6 \\
                \cmidrule(lr){2-8}
                & FADE (ours) & \textbf{96.3$\pm$\small{0.4}} & \textbf{98.1$\pm$\small{0.2}} & \textbf{95.5$\pm$\small{0.4}} & \textbf{90.7$\pm$\small{0.3}} & \textbf{91.9$\pm$\small{0.4}} & \textbf{87.0$\pm$\small{0.2}} \\
                \bottomrule
            \end{tabular}
        }
        \caption{Comparison of AC performance on MVTec-AD and VisA. We report the mean and standard deviation over 5 random seeds. Bold indicates the best performance.}
        \label{tab:main_ac}
    \end{minipage}    
    \hfill
    \begin{minipage}[c]{0.295\textwidth}
        \centering
        \resizebox{\textwidth}{!}{%
            \begin{tabular}{cc|l}
            \toprule
            Task & Shot & Aggregation \\
            \midrule
            AC & 0-shot & $s^{lang}$  \\
            & k-shot & $s^{lang}+s^{vis,k}$  \\
            \midrule
            AS & 0-shot & $\boldsymbol{M}^{lang}+\boldsymbol{M}^{vis,0}$  \\
            & k-shot & $\boldsymbol{M}^{lang}+\boldsymbol{M}^{vis,k}$  \\
            \bottomrule
            \end{tabular}
        }
        \caption{Language and vision aggregation.}
        \label{tab:lang_vis_aggregation}
        
        \vfill
        \centering
        \resizebox{\textwidth}{!}{%
            \begin{tabular}{l|cc}
                \toprule
                \makecell[cl]{Patch \\ embeddings} & MVTec-AD & VisA \\
                \midrule
                CLIP & 18.0 & 13.9  \\
                GEM & \textbf{86.5}  & \textbf{87.0} \\
                \bottomrule
            \end{tabular}
        }
        \caption{CLIP vs GEM embeddings for 0-shot AS (pAUROC).}
        \label{tab:ablation_clip_vs_gem_as}
    \end{minipage}
\end{table}

We conduct a series of experiments to evaluate the performance of FADE on AC and AS under the zero-/few-shot regime. We also perform ablation experiments to study the impact of each component in FADE.

\mysubsubsection{Datasets}
All experiments are based on the MVTec-AD~\cite{bergmann2019mvtecad} and VisA~\cite{zou2022visa} datasets which are standard benchmarks for AC and AS. The datasets include a range of different objects such as capsule and cashew. We only use the test split for evaluation which contains both normal and anomalous images and their ground truth segmentation masks. The training split is only used for sampling the reference images during few-shot evaluation.

\mysubsubsection{Evaluation metrics}
In accordance with prior works~\cite{roth2022patchcore, jeong2023winclip}, we report the following evaluation metrics. For AC, we report Area Under the Receiver Operating Characteristics curve (AUROC), Area Under the Precision-Recall curve (AUPR) and F1-score at optimal threshold (F1-max). For AS, we report pixel-wise AUROC (pAUROC), Per-Region Overlap (PRO) and pixel-wise F1-max.

\mysubsubsection{Implementation details}
For CLIP, we use the OpenCLIP\footnote{https://github.com/mlfoundations/open\_clip} implementation of ViT-B/16+ ($240\times 240$) trained on LAION-400M \cite{schuhmann2021laion}. For GEM, we use its official implementation\footnote{https://github.com/WalBouss/GEM}.

\subsection{Zero-/Few-Shot Anomaly Classification and Segmentation}

Tab.~\ref{tab:main_ac} compares the AC performance of FADE with two prior work, PatchCore~\cite{roth2022patchcore} which is an unsupervised anomaly detection method and WinCLIP~\cite{jeong2023winclip} which is a state-of-the-art zero-/few-shot anomaly detection method. For the zero-shot setting, FADE is a reproduction of WinCLIP without any new additions. For the few-shot setting, FADE outperforms the other methods on all metrics for both MVTec-AD and VisA with a larger improvement seen on VisA. This shows that the vision-guided AS component of FADE is also beneficial to AC.

Tab.~\ref{tab:main_as} compares the AS performance of FADE with PatchCore and WinCLIP. Under zero-shot, FADE significantly outperforms other methods on all metrics for both MVTec-AD and VisA. This is due to the various AS improvements of FADE which include the multi-scale GEM embeddings, a better ChatGPT prompt ensemble and the zero-shot vision-guidance using query image. Under few-shot, FADE performs similar to WinCLIP on MVTec-AD and again outperforms WinCLIP on VisA which is the more challenging dataset. This shows the advantage of FADE on more difficult images. It also shows that vanilla CLIP patch embeddings can be used in place of WinCLIP embeddings for few-shot vision-guided AS. Furthermore, FADE performance has a lower standard deviations across different random runs. This indicates that FADE is more robust to the selection of different reference images. Fig.~\ref{fig:qualitative_main} shows qualitative AS results for some objects. See supplementary for more.

Additional quantitative results comparing FADE and other state-of-the-art methods such as AnomalyCLIP~\cite{zhou2024anomalyclip}, AnomalyGPT~\cite{gu2024anomalygpt} and APRIL-GAN~\cite{chen2023aprilgan} can be found in the supplementary. FADE performs competitively even though these other methods require training using additional anomaly detection datasets while FADE does not need any further training.

\subsection{Ablation Study}
\begin{table}[t]
    \centering
    \begin{minipage}[c]{0.709\textwidth}
        \centering
        \resizebox{\textwidth}{!}{%
            \begin{tabular}{clcccccc}
                \toprule
                \multicolumn{2}{l}{Anomaly Segmentation} & \multicolumn{3}{c}{MVTec-AD} & \multicolumn{3}{c}{VisA} \\
                \cmidrule(lr){3-5} \cmidrule(lr){6-8}
                Setup & Method & pAUROC & PRO & $F1$-max & pAUROC & PRO & $F1$-max\\
                \midrule
                0-shot & WinCLIP & 85.1$\pm$\small{0.0} & 64.6$\pm$\small{0.0} & 31.7$\pm$\small{0.0} & 79.6$\pm$\small{0.0} & 56.8$\pm$\small{0.0} & 14.8$\pm$\small{0.0} \\
                \cmidrule(lr){2-8}
                & FADE (ours) & \textbf{89.6$\pm$\small{0.0}} & \textbf{84.5$\pm$\small{0.0}} & \textbf{39.8$\pm$\small{0.0}} & \textbf{91.5$\pm$\small{0.0}} & \textbf{79.3$\pm$\small{0.0}} & \textbf{16.7$\pm$\small{0.0}} \\
                \midrule
                1-shot & PatchCore & 92.0$\pm$\small{1.0} & 79.7$\pm$\small{2.0} & 50.4$\pm$\small{2.1} & 95.4$\pm$\small{0.6} & 80.5$\pm$\small{2.5} & 38.0$\pm$\small{1.9} \\
                & WinCLIP+ & 95.2$\pm$\small{0.5} & 87.1$\pm$\small{1.2} & \textbf{55.9$\pm$\small{2.7}} & 96.4$\pm$\small{0.4} & 85.1$\pm$\small{2.1} & 41.3$\pm$\small{2.3} \\
                \cmidrule(lr){2-8}
                & FADE (ours) & \textbf{95.4$\pm$\small{0.3}} & \textbf{88.3$\pm$\small{0.3}} & 54.6$\pm$\small{1.1} & \textbf{97.5$\pm$\small{0.1}} & \textbf{88.9$\pm$\small{0.7}} & \textbf{42.3$\pm$\small{0.5}} \\
                \midrule
                2-shot & PatchCore & 93.3$\pm$\small{0.6} & 82.3$\pm$\small{1.3} & 53.0$\pm$\small{1.7} & 96.1$\pm$\small{0.5} & 82.6$\pm$\small{2.3} & 41.0$\pm$\small{3.9} \\
                & WinCLIP+ & \textbf{96.0$\pm$\small{0.3}} & 88.4$\pm$\small{0.9} & \textbf{58.4$\pm$\small{1.7}} & 96.8$\pm$\small{0.3} & 86.2$\pm$\small{1.4} & 43.5$\pm$\small{3.3} \\
                \cmidrule(lr){2-8}
                & FADE (ours) & 95.8$\pm$\small{0.2} & \textbf{88.9$\pm$\small{0.2}} & 55.8$\pm$\small{1.0} & \textbf{97.8$\pm$\small{0.1}} & \textbf{89.8$\pm$\small{0.4}} & \textbf{44.4$\pm$\small{0.9}} \\
                \midrule
                4-shot & PatchCore & 94.3$\pm$\small{0.5} & 84.3$\pm$\small{1.6} & 55.0$\pm$\small{1.9} & 96.8$\pm$\small{0.3} & 84.9$\pm$\small{1.4} & 43.9$\pm$\small{3.1} \\
                & WinCLIP+ & \textbf{96.2$\pm$\small{0.3}} & 89.0$\pm$\small{0.8} & \textbf{59.5$\pm$\small{1.8}} & 97.2$\pm$\small{0.2} & 87.6$\pm$\small{0.9} & \textbf{47.0$\pm$\small{3.0}} \\
                \cmidrule(lr){2-8}
                & FADE (ours) & \textbf{96.2$\pm$\small{0.1}} & \textbf{89.5$\pm$\small{0.2}} & 57.0$\pm$\small{0.8} & \textbf{98.0$\pm$\small{0.0}} & \textbf{90.0$\pm$\small{0.4}} & 46.4$\pm$\small{0.7} \\
                \bottomrule
            \end{tabular}
        }
        \caption{
        Comparison of AS performance on MVTec-AD and VisA. We report the mean and standard deviation over 5 random seeds. Bold indicates the best performance.
        }
        \label{tab:main_as}
    \end{minipage}    
    \hfill
    \begin{minipage}[c]{0.281\textwidth}
        \centering
        \resizebox{\textwidth}{!}{%
            \begin{tabular}{cccc}
                \toprule
                \multicolumn{2}{c}{AC (AUROC)} & \multicolumn{2}{c}{\# shots} \\
                \cmidrule(r){1-2} \cmidrule(l){3-4}
                $s^{lang}$ & $s^{vis}$ & 1     & 4 \\
                \cmidrule(r){1-2} \cmidrule(l){3-4}
                \ding{51} & \ding{55} & 90.0  & 90.0 \\
                \ding{55} & \ding{51} & 90.7  & 94.5 \\
                \cmidrule(r){1-2} \cmidrule(l){3-4}
                \ding{51} & \ding{51} & \textbf{93.9}  & \textbf{96.3} \\
                \bottomrule
            \end{tabular}
        }
        \caption{Language- vs vision-guided AC.}
        \label{tab:ablation_lang_vs_vis_mvtec_ac}
        
        \vfill
        \centering
        \resizebox{\textwidth}{!}{%
            \begin{tabular}{ccccc}
                \toprule
                \multicolumn{2}{c}{AS (pAUROC)} & \multicolumn{3}{c}{\# shots} \\
                \cmidrule(r){1-2} \cmidrule(l){3-5}
                $\boldsymbol{M}^{lang}$ & $\boldsymbol{M}^{vis}$ & 0 & 1     & 4 \\
                \cmidrule(r){1-2} \cmidrule(l){3-5}
                \ding{51} & \ding{55} & 86.5 & 86.5  & 86.5 \\
                \ding{55} & \ding{51} & 86.6 & 95.1  & 96.1 \\
                \cmidrule(r){1-2} \cmidrule(l){3-5}
                \ding{51} & \ding{51} & \textbf{89.6} & \textbf{95.4}  & \textbf{96.2} \\
                \bottomrule
            \end{tabular}
        }
        \caption{Language- vs vision-guided AS.}
        \label{tab:ablation_lang_vs_vis_mvtec_as}
    \end{minipage}
\end{table}

\mysubsubsection{GEM for zero-shot AS}
Tab.~\ref{tab:ablation_clip_vs_gem_as} shows the impact of using CLIP vs GEM patch embeddings for zero-shot language-guided AS. When CLIP embeddings are used, pAUROC is below 50.0 due to the problem of opposite visualisation between the normal and anomalous regions. See supplementary for qualitative examples of the opposite visualisation when CLIP embeddings are used while GEM embeddings fix this problem. More result comparison on CLIP vs GEM embeddings for the different components of FADE are also shown in the supplementary. The result motivates the choice of CLIP or GEM embeddings used in Fig.~\ref{fig:method}a-d.

\mysubsubsection{Prompt ensemble for zero-shot AS}
Tab.~\ref{tab:ablation_prompts} shows the improvement on zero-shot language-guided AS when we use additional text prompts generated automatically by ChatGPT. These prompts capture more diverse descriptions related to the concept of anomaly and normality which allow us to better use the CLIP model for improved AS. Interestingly, we note that the ChatGPT prompts did not improve the zero-shot image-level AC performance.

\mysubsubsection{Multi-scale aggregation}
Tab.~\ref{tab:ablation_img_scale_mvtec_ac} and \ref{tab:ablation_img_scale_mvtec_as} show the AC and AS performance on MVTec-AD when different input image scale/size is used. Since the patch size is constant (16 pixels), a larger input size means that each patch is covering a finer scale and anomaly with smaller size will be detected and vice versa. Multi-scale refers to taking the average outputs from the three image scales and it has the best results at all k-shot settings since it allows for detecting anomalies of different sizes. See supplementary for qualitative results.

\mysubsubsection{Language vs vision guidance}
Tab.~\ref{tab:ablation_lang_vs_vis_mvtec_ac} and \ref{tab:ablation_lang_vs_vis_mvtec_as} show the impact of language- vs vision-guided AC and AS on MVTec-AD. Language guidance depends on text prompts while vision guidance depends on the query (0-shot) or reference images (k-shot). The two types of guidance are complementary and improve the overall performance. See supplementary for qualitative results.

\begin{table}[t]
    \centering
    \begin{minipage}[b]{0.4\textwidth}
        \centering
        \resizebox{\textwidth}{!}{%
            \begin{tabular}{l|cc}
                \toprule
                Prompt ensemble & MVTec-AD & VisA \\
                \midrule
                WinCLIP prompts & 83.7 & 74.0  \\
                \midrule
                \makecell[l]{WinCLIP + \\ ChatGPT prompts} & \textbf{86.5}  & \textbf{87.0} \\
                \bottomrule
            \end{tabular}
        }
        \caption{Prompt ablations on language-guided AS (pAUROC).}
        \label{tab:ablation_prompts}
    \end{minipage}    
    \hfill
    \begin{minipage}[b]{0.258\textwidth}
        \centering
        \resizebox{\textwidth}{!}{%
            \begin{tabular}{cccc}
                \toprule
                AC (AUROC) & \multicolumn{3}{c}{\# shots} \\
                \cmidrule(l){2-4}
                Image scale & 1     & 2     & 4 \\
                \cmidrule(r){1-1} \cmidrule(l){2-4}
                240 & 92.6  & 94.4  & 95.3 \\
                448 & 92.3  & 93.7  & 94.6 \\
                896 & 90.5  & 91.5  & 92.7 \\
                \cmidrule(r){1-1} \cmidrule(l){2-4}
                Multi-scale & \textbf{93.9}  & \textbf{95.2}  & \textbf{96.3} \\
                \bottomrule
            \end{tabular}
        }
        \caption{Image scale ablations for AC.}
        \label{tab:ablation_img_scale_mvtec_ac}
    \end{minipage}
    \hfill
    \begin{minipage}[b]{0.313\textwidth}
        \centering
        \resizebox{\textwidth}{!}{%
            \begin{tabular}{ccccc}
                \toprule
                AS (pAUROC) & \multicolumn{4}{c}{\# shots} \\
                \cmidrule(l){2-5}
                Image scale & 0     & 1     & 2     & 4 \\
                \cmidrule(r){1-1} \cmidrule(l){2-5}
                240 & 87.6  & 93.4  & 94.0  & 94.5 \\
                448 & 87.4  & 93.5  & 94.0  & 94.5 \\
                896 & 84.7  & 91.3  & 91.7  & 92.1 \\
                \cmidrule(r){1-1} \cmidrule(l){2-5}
                Multi-scale & \textbf{89.6}  & \textbf{95.4}  & \textbf{95.8}  & \textbf{96.2} \\
                \bottomrule
            \end{tabular}
        }
        \caption{Image scale ablations for AS.}
        \label{tab:ablation_img_scale_mvtec_as}
    \end{minipage}
\end{table}
\section{Conclusion and Future Work}
\label{sec:conclusion}

We present a new method FADE that utilises and adapts the CLIP model for zero-/few-shot anomaly detection guided by language and vision. First, we improve language-guided anomaly segmentation by using multi-scale GEM embeddings that are better aligned with language than CLIP embeddings. This is further improved by using an LLM to generate a new prompt ensemble that better captures the concept of anomaly and normality. Finally, vision guidance from the query image further boosts the zero-shot anomaly detection performance while vision guidance from the reference images extends the method to the few-shot setting. On standard benchmarks, FADE performs competitively compared to other state-of-the-art methods with the largest margin of improvement on zero-shot anomaly segmentation.

There are several points about FADE that are worth investigating as future work. First, the use of text prompts generated by ChatGPT is not reproducible. While an alternative open-source LLM can be used to set the seed to ensure reproducibility, it is also worth investigating how sensitive the anomaly detection performance is to the different prompts generated by the same LLM or across different LLMs.

Our experiment results show that for language-guided anomaly detection, GEM embeddings are better for zero-shot AS while CLIP embeddings are more suited for zero-shot AC. On the other hand, for vision-guided anomaly detection, GEM embeddings are better for zero-shot AS while CLIP embeddings are more suited for few-shot AS. A more comprehensive study is needed to gain a better understanding of when to use the GEM and CLIP embeddings under the different detection pipelines and scenarios.

The vision-guided zero-shot AS constructs the memory bank from its own query image patches. This works well at detecting anomalies in textural images (E.g.\ leather) that normally contain visually similar patches. However, this works less well and produces false positives for anomaly detection in object images (E.g.\ transistor) whose patches differ to a greater extent. Future work needs to explore other vision-guided zero-shot methods to address this issue. 
\section*{Acknowledgement}
\label{sec:acknowledgement}

We would like to thank Romain Sabathe for the insightful discussion and Olivier Koch for the full support throughout this project.
\bibliography{egbib}

\begin{thebibliography}{25}
\providecommand{\natexlab}[1]{#1}
\providecommand{\url}[1]{\texttt{#1}}
\expandafter\ifx\csname urlstyle\endcsname\relax
  \providecommand{\doi}[1]{doi: #1}\else
  \providecommand{\doi}{doi: \begingroup \urlstyle{rm}\Url}\fi

\bibitem[Bergmann et~al.(2019)Bergmann, Fauser, Sattlegger, and Steger]{bergmann2019mvtecad}
Paul Bergmann, Michael Fauser, David Sattlegger, and Carsten Steger.
\newblock {MVTec-AD} — a comprehensive real-world dataset for unsupervised anomaly detection.
\newblock In \emph{2019 IEEE/CVF Conference on Computer Vision and Pattern Recognition (CVPR)}, pages 9584--9592, 2019.
\newblock \doi{10.1109/CVPR.2019.00982}.

\bibitem[Bergmann et~al.(2020)Bergmann, Fauser, Sattlegger, and Steger]{bergmann2020studentteacher}
Paul Bergmann, Michael Fauser, David Sattlegger, and Carsten Steger.
\newblock Uninformed students: Student-teacher anomaly detection with discriminative latent embeddings.
\newblock In \emph{Proceedings of the IEEE/CVF conference on computer vision and pattern recognition}, pages 4183--4192, 2020.

\bibitem[Bousselham et~al.(2023)Bousselham, Petersen, Ferrari, and Kuehne]{bousselham2023gem}
Walid Bousselham, Felix Petersen, Vittorio Ferrari, and Hilde Kuehne.
\newblock Grounding everything: Emerging localization properties in vision-language transformers, 2023.

\bibitem[Cao et~al.(2023)Cao, Xu, Sun, Cheng, Du, Gao, and Shen]{cao2023saa}
Yunkang Cao, Xiaohao Xu, Chen Sun, Yuqi Cheng, Zongwei Du, Liang Gao, and Weiming Shen.
\newblock Segment any anomaly without training via hybrid prompt regularization, 2023.

\bibitem[Chen et~al.(2023)Chen, Han, and Zhang]{chen2023aprilgan}
Xuhai Chen, Yue Han, and Jiangning Zhang.
\newblock A zero-/few-shot anomaly classification and segmentation method for {CVPR 2023 VAND} workshop challenge tracks 1\&2: 1st place on zero-shot {AD} and 4th place on few-shot {AD}.
\newblock \emph{arXiv preprint arXiv:2305.17382}, 2023.

\bibitem[Defard et~al.(2021)Defard, Setkov, Loesch, and Audigier]{defard2021padim}
Thomas Defard, Aleksandr Setkov, Angelique Loesch, and Romaric Audigier.
\newblock {PaDiM}: a patch distribution modeling framework for anomaly detection and localization.
\newblock In \emph{International Conference on Pattern Recognition}, pages 475--489. Springer, 2021.

\bibitem[Gu et~al.(2024)Gu, Zhu, Zhu, Chen, Tang, and Wang]{gu2024anomalygpt}
Zhaopeng Gu, Bingke Zhu, Guibo Zhu, Yingying Chen, Ming Tang, and Jinqiao Wang.
\newblock Anomalygpt: Detecting industrial anomalies using large vision-language models.
\newblock In \emph{Proceedings of the AAAI Conference on Artificial Intelligence}, volume~38, pages 1932--1940, 2024.

\bibitem[Gudovskiy et~al.(2022)Gudovskiy, Ishizaka, and Kozuka]{gudovskiy2022cflow}
Denis Gudovskiy, Shun Ishizaka, and Kazuki Kozuka.
\newblock {CFLOW-AD}: Real-time unsupervised anomaly detection with localization via conditional normalizing flows.
\newblock In \emph{Proceedings of the IEEE/CVF winter conference on applications of computer vision}, pages 98--107, 2022.

\bibitem[Huang et~al.(2022)Huang, Guan, Jiang, Zhang, Spratling, and Wang]{huang2022regad}
Chaoqin Huang, Haoyan Guan, Aofan Jiang, Ya~Zhang, Michael Spratling, and Yan-Feng Wang.
\newblock Registration based few-shot anomaly detection.
\newblock In \emph{European Conference on Computer Vision}, pages 303--319. Springer, 2022.

\bibitem[Jeong et~al.(2023)Jeong, Zou, Kim, Zhang, Ravichandran, and Dabeer]{jeong2023winclip}
Jongheon Jeong, Yang Zou, Taewan Kim, Dongqing Zhang, Avinash Ravichandran, and Onkar Dabeer.
\newblock {WinCLIP}: Zero-/few-shot anomaly classification and segmentation.
\newblock In \emph{2023 IEEE/CVF Conference on Computer Vision and Pattern Recognition (CVPR)}, pages 19606--19616, 2023.
\newblock \doi{10.1109/CVPR52729.2023.01878}.

\bibitem[Kirillov et~al.(2023)Kirillov, Mintun, Ravi, Mao, Rolland, Gustafson, Xiao, Whitehead, Berg, Lo, et~al.]{kirillov2023sam}
Alexander Kirillov, Eric Mintun, Nikhila Ravi, Hanzi Mao, Chloe Rolland, Laura Gustafson, Tete Xiao, Spencer Whitehead, Alexander~C Berg, Wan-Yen Lo, et~al.
\newblock Segment anything.
\newblock In \emph{Proceedings of the IEEE/CVF International Conference on Computer Vision}, pages 4015--4026, 2023.

\bibitem[Li et~al.(2021)Li, Sohn, Yoon, and Pfister]{li2021cutpaste}
Chun-Liang Li, Kihyuk Sohn, Jinsung Yoon, and Tomas Pfister.
\newblock {CutPaste}: Self-supervised learning for anomaly detection and localization.
\newblock In \emph{Proceedings of the IEEE/CVF conference on computer vision and pattern recognition}, pages 9664--9674, 2021.

\bibitem[Li et~al.(2024{\natexlab{a}})Li, Cao, Ye, Ding, Tu, and Chen]{li2024clipsam}
Shengze Li, Jianjian Cao, Peng Ye, Yuhan Ding, Chongjun Tu, and Tao Chen.
\newblock {ClipSAM}: {CLIP} and {SAM} collaboration for zero-shot anomaly segmentation, 2024{\natexlab{a}}.

\bibitem[Li et~al.(2024{\natexlab{b}})Li, Zhang, Tan, Chen, Qu, Xie, and Ma]{li2024promptad}
Xiaofan Li, Zhizhong Zhang, Xin Tan, Chengwei Chen, Yanyun Qu, Yuan Xie, and Lizhuang Ma.
\newblock {PromptAD}: Learning prompts with only normal samples for few-shot anomaly detection, 2024{\natexlab{b}}.

\bibitem[Li et~al.(2023)Li, Wang, Duan, and Li]{li2023clipsurgery}
Yi~Li, Hualiang Wang, Yiqun Duan, and Xiaomeng Li.
\newblock {CLIP} surgery for better explainability with enhancement in open-vocabulary tasks, 2023.

\bibitem[Liu et~al.(2023)Liu, Zeng, Ren, Li, Zhang, Yang, Li, Yang, Su, Zhu, et~al.]{liu2023groundingdino}
Shilong Liu, Zhaoyang Zeng, Tianhe Ren, Feng Li, Hao Zhang, Jie Yang, Chunyuan Li, Jianwei Yang, Hang Su, Jun Zhu, et~al.
\newblock Grounding {DINO}: Marrying dino with grounded pre-training for open-set object detection.
\newblock \emph{arXiv preprint arXiv:2303.05499}, 2023.

\bibitem[Radford et~al.(2021)Radford, Kim, Hallacy, Ramesh, Goh, Agarwal, Sastry, Askell, Mishkin, Clark, et~al.]{radford2021clip}
Alec Radford, Jong~Wook Kim, Chris Hallacy, Aditya Ramesh, Gabriel Goh, Sandhini Agarwal, Girish Sastry, Amanda Askell, Pamela Mishkin, Jack Clark, et~al.
\newblock Learning transferable visual models from natural language supervision.
\newblock In \emph{International conference on machine learning}, pages 8748--8763. PMLR, 2021.

\bibitem[Roth et~al.(2022)Roth, Pemula, Zepeda, Sch{\"o}lkopf, Brox, and Gehler]{roth2022patchcore}
Karsten Roth, Latha Pemula, Joaquin Zepeda, Bernhard Sch{\"o}lkopf, Thomas Brox, and Peter Gehler.
\newblock Towards total recall in industrial anomaly detection.
\newblock In \emph{Proceedings of the IEEE/CVF Conference on Computer Vision and Pattern Recognition}, pages 14318--14328, 2022.

\bibitem[Rudolph et~al.(2021)Rudolph, Wandt, and Rosenhahn]{rudolph2021differnet}
Marco Rudolph, Bastian Wandt, and Bodo Rosenhahn.
\newblock Same same but differnet: Semi-supervised defect detection with normalizing flows.
\newblock In \emph{Proceedings of the IEEE/CVF winter conference on applications of computer vision}, pages 1907--1916, 2021.

\bibitem[Schuhmann et~al.(2021)Schuhmann, Vencu, Beaumont, Kaczmarczyk, Mullis, Katta, Coombes, Jitsev, and Komatsuzaki]{schuhmann2021laion}
Christoph Schuhmann, Richard Vencu, Romain Beaumont, Robert Kaczmarczyk, Clayton Mullis, Aarush Katta, Theo Coombes, Jenia Jitsev, and Aran Komatsuzaki.
\newblock {LAION-400m}: Open dataset of {CLIP}-filtered 400 million image-text pairs.
\newblock \emph{arXiv preprint arXiv:2111.02114}, 2021.

\bibitem[Tamura(2023)]{tamura2023randomword}
Masato Tamura.
\newblock Random word data augmentation with {CLIP} for zero-shot anomaly detection.
\newblock In \emph{34th British Machine Vision Conference 2023, {BMVC} 2023, Aberdeen, UK, November 20-24, 2023}. BMVA, 2023.
\newblock URL \url{https://papers.bmvc2023.org/0018.pdf}.

\bibitem[Wu et~al.(2021)Wu, Chen, Fuh, and Liu]{wu2021metaformer}
Jhih-Ciang Wu, Ding-Jie Chen, Chiou-Shann Fuh, and Tyng-Luh Liu.
\newblock Learning unsupervised metaformer for anomaly detection.
\newblock In \emph{Proceedings of the IEEE/CVF International Conference on Computer Vision}, pages 4369--4378, 2021.

\bibitem[Zhou et~al.(2022)Zhou, Loy, and Dai]{zhou2022maskclip}
Chong Zhou, Chen~Change Loy, and Bo~Dai.
\newblock Extract free dense labels from {CLIP}.
\newblock In \emph{European Conference on Computer Vision}, pages 696--712. Springer, 2022.

\bibitem[Zhou et~al.(2024)Zhou, Pang, Tian, He, and Chen]{zhou2024anomalyclip}
Qihang Zhou, Guansong Pang, Yu~Tian, Shibo He, and Jiming Chen.
\newblock Anomaly{CLIP}: Object-agnostic prompt learning for zero-shot anomaly detection.
\newblock In \emph{The Twelfth International Conference on Learning Representations}, 2024.
\newblock URL \url{https://openreview.net/forum?id=buC4E91xZE}.

\bibitem[Zou et~al.(2022)Zou, Jeong, Pemula, Zhang, and Dabeer]{zou2022visa}
Yang Zou, Jongheon Jeong, Latha Pemula, Dongqing Zhang, and Onkar Dabeer.
\newblock {SPot-the-Difference} self-supervised pre-training for anomaly detection and segmentation.
\newblock In \emph{European Conference on Computer Vision}, pages 392--408. Springer, 2022.

\end{thebibliography}
\end{document}